\documentclass[10pt,twocolumn,letterpaper]{article}

\usepackage{iccv}
\usepackage{times}
\usepackage{epsfig}
\usepackage{graphicx}
\usepackage{amsmath}
\usepackage{amssymb}
\usepackage{booktabs}
\usepackage{flushend}

\graphicspath{{images/}}

\iccvfinalcopy 

\def\iccvPaperID{20} 

\ificcvfinal\fi
\begin{document}

\title{ARIGAN: Synthetic Arabidopsis Plants using Generative Adversarial Network}

\author{Mario Valerio Giuffrida\\
The Alan Turing Institute\\
92 Euston Road, London\\
{\tt\small vgiuffrida@turing.ac.uk}
\and
Hanno Scharr\\
Forschungszentrum J\"ulich\\
IBG-2, J\"ulich\\
{\tt\small h.scharr@fz-juelich.de}
\and
Sotirios A Tsaftaris\\
University of Edinburgh\\
AGB, King's Building, Edinburgh\\
{\tt\small S.Tsaftaris@ed.ac.uk}
}

\maketitle

\begin{abstract}

In  recent years, there has been an increasing interest in image-based plant phenotyping, applying state-of-the-art machine learning approaches to tackle challenging problems, such as leaf segmentation (a multi-instance problem) and counting. Most of these algorithms need labelled data to learn a model for the task at hand. Despite the recent release of a few plant phenotyping datasets, large annotated plant image datasets for the purpose of training deep learning algorithms are lacking. One common approach to alleviate the lack of training data is \textit{dataset augmentation}. Herein, we propose an alternative solution to dataset augmentation for plant phenotyping, creating artificial images of plants using generative neural networks. We propose the {\normalfont Arabidopsis Rosette Image Generator (through) Adversarial Network}: a deep convolutional network that is able to generate synthetic rosette-shaped plants, inspired by DCGAN (a recent adversarial network model using convolutional layers). Specifically, we trained the network using A1, A2, and A4 of the CVPPP 2017 LCC dataset, containing Arabidopsis Thaliana plants. We show that our model is able to generate realistic $128\times128$ colour images of plants. We train our network conditioning on leaf count, such that it is possible to generate plants with a given number of leaves suitable, among others, for training regression based models. We propose a new Ax dataset of artificial plants images, obtained by our ARIGAN. We evaluate this new dataset using a state-of-the-art leaf counting algorithm, showing that the testing error is reduced when Ax is used as part of the training data.

\end{abstract}

\section{Introduction}

It is widely known that machine learning has brought a massive benefit to many areas over the last decades. In many scenarios, computer vision relies on vast amount of data to train machine learning algorithms. Recently, several datasets of top-view plant images were publicly released \cite{Bell2016,Cruz2016,Minervini2016}, allowing the computer vision community to propose new methodologies for leaf counting \cite{Giuffrida2015,Minervini2017, Pape2015} and leaf segmentation \cite{Ren2016,RomeraParedes2016,Scharr2016}. However, these algorithms can perform even better when provided with more training data, such that the generalisation capabilities of the trained model are increased, while also reducing overfitting.

As of now, a main issue affecting current plant phenotyping datasets is the limited quantity of labelled data \cite{Tsaftaris2016}. 
Typically, the computer vision community  has been employing \textit{dataset augmentation} to increase the amount of data using artificial transformations. In fact, artificially perturbing the original dataset with affine transformations (e.g., rotation, scale, translation) is considered a common practice. However, this approach has limits: the augmented data \textit{only} capture the variability of the training set (e.g., if a plant with 7 leaves is missing from the training set, this particular instance will not ever be learnt). For this reason, herein we present preliminary work on generating artificial images of plants, using a recent generative model that learns how to create new images.

\begin{figure}
\includegraphics[width=\linewidth]{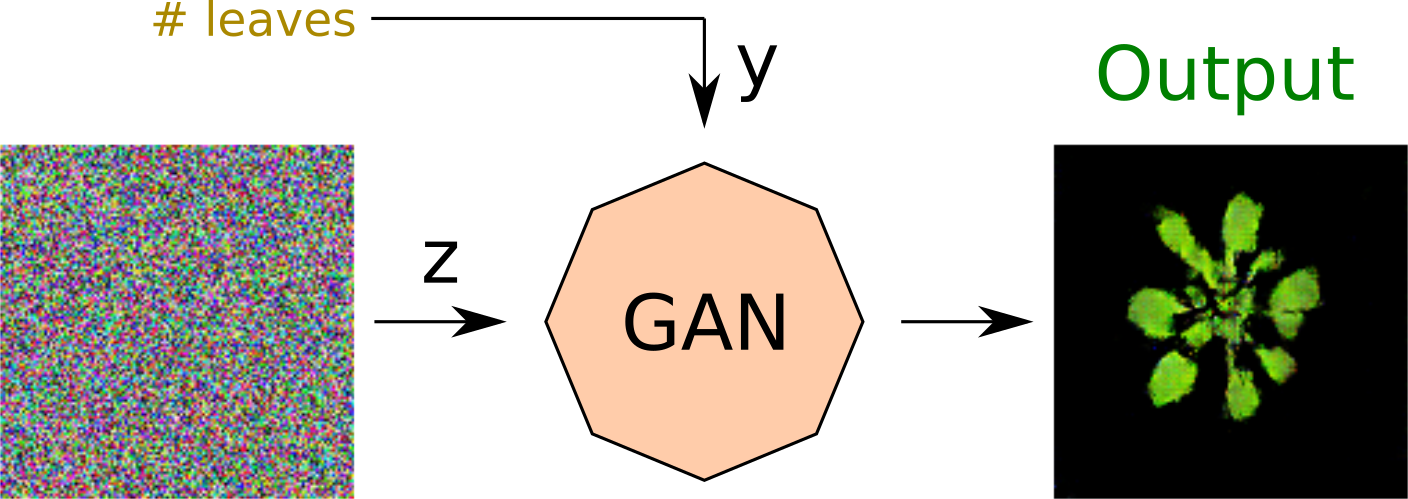}
\caption{Schematic of the proposed method: a conditional generative adversarial network is trained to map random uniform noise $z$ into Arabidopsis plants, given a condition $y$ on the number of leaves to generate.}
\label{fig:proposed}
\end{figure}

\begin{figure*}[t]
\includegraphics[width=0.9\textwidth]{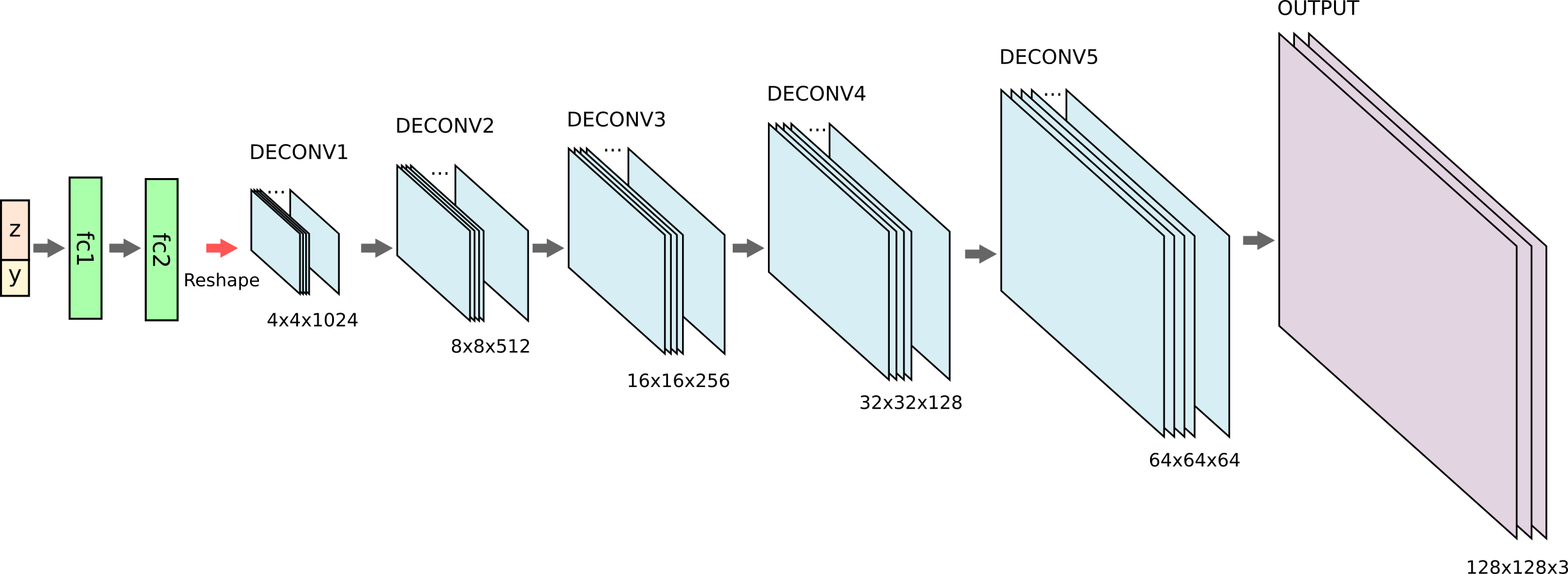}
\caption{\textbf{Generator.} This network takes as input a variable $z$ (random noise) together with the condition vector $y$. These inputs are then provided to two fully connected layers, where \textit{fc2} has the same amount of hidden units of the first deconvolutional layer. The information is then processed through 5 deconvolutional layers, where the last one provides an 128x128 RGB image. The condition $y$ is applied to all the stages (fc and deconv layers). We showed it in the input only for sake of clarity.}
\label{fig:G}
\end{figure*}

An attempt to generate rosettes was done in \cite{Muendermann2005}, where an empirical model was created  analysing 5 Arabidopsis plants of about 11 leaves. Sigmoidal growth models were fitted based on the sets of plants under their study. Then, organs were dissected and leaves were used to fit B-splines to obtain vector images. The issues of this model can be summarised as follows: (i) the approach is confined on the manual observation of morphological traits of a limited set of plants; and (ii) there is a lack of realism in the generated images, for example in the absence of texture on the leaves.

The manual acquisition of morphological data from plants is a tedious and error-prone process \cite{Giuffrida2015,Minervini2017} and we aim to alleviate this process, by using neural networks that can learn those parameters from data. Recently, several generative models were proposed to generate realistic images. In the literature it is possible to find different generative models to create artificial images. For example, in \cite{Cappelli2000} the authors synthesise images of fingerprints, reproducing the orientation model of fingerprint lines. Another method to generate images employs genetic programming \cite{DiPaola2009}. However, recent interest in neural networks, has brought new methodologies to generate synthetic images. In fact, convolutional neural networks (CNNs) were used to generate images of photorealistic chairs \cite{Dosovitskiy2015}. In \cite{Gregor2015}, the authors introduce the \textit{Deep Recurrent Attentive Writer} (DRAW) network, which combines LSTM \cite{Hochreiter1997} layers to draw images, using a selective attention model that, at each iteration, finds the next location to draw new pixels. Despite its impressive results, this method is challenged by natural image data.  The \textit{Generative Adversarial Network} (GAN) \cite{Goodfellow2014} has been proven to be successful at generating synthetic images also on natural images. In a GAN, there are two models competing with each other: the \textit{Generator} (G), which creates artificial images; and the \textit{Discriminator} (D), which is trained to classify images coming from the training set (real) and the generator (fake). The spirit of the GAN is to improve G to create more realistic images, whilst D is trained to distinguish between real and generated images. Training works by improving in alternating fashion G or D, until an equilibrium is obtained. Generally speaking, the generator and the discriminator can be any network that satisfies the following criteria: (i) D needs to take as input an image and has to output `1' and `0' (real/not real); (ii) G needs to take as input random noise (e.g. drawn from an uniform or normal distribution) and has to give as output an image. LAPGAN \cite{Denton2015} was proposed, which was able to produce better quality images using Laplacian pyramids. A new successful adversarial network providing outstanding results is \textit{Deep Convolutional GAN} \cite{Radford2015}. The benefits of this model mostly stem from the use of convolutional/deconvolutional layers for discriminator and generator respectively and the lack of pooling/upsampling layers. 

Although adversarial networks have brought many benefits, a main limitation is the lack of direct control over the images generated. For instance, in the case where we want to train a GAN to generate images of handwritten digits --the MNIST dataset \cite{mnist} is a typical benchmark dataset in computer vision and machine learning--, it would be reasonable to have control over which digit to generate each time. For this reason \textit{Conditional GAN} \cite{Mirza2014} was proposed to overcome such limitation. In this new formulation, generator and discriminator networks are endowed with an additional input, allowing to be trained under certain conditions. In \cite{Han2016}, the authors propose \textit{StackGAN}, a two-stage GAN conditioned on image captions. Specifically, Stage-I generates coarse images, which are fed to Stage-II to obtain more realistic images. 


In this paper, we show how to generate Arabidopsis plants, using a model inspired by \cite{Radford2015}, trained on the CVPPP 2017\footnote{Available at the following URL: \url{https://www.plant-phenotyping.org/CVPPP2017}}  dataset. The network learns how to map  random noise $z$ into an \textit{Arabidopsis} plant, under a condition $y$.  For our purposes, $y$ encodes the number of leaves that the artificially generated plant should have. The employed model, which we call \textit{Arabidopsis Rosette Image Generator (through) Adversarial Network} (ARIGAN), is able to create $128\times 128$ RGB images of Arabidopsis plants, as shown in \figurename~\ref{fig:proposed}. We evaluate our model by creating an \textit{Ax} (using the CVPPP dataset name convention) and provide the generated data to a state-of-the-art leaf counting algorithm \cite{Giuffrida2015} to augment the training dataset.

The remainder of this paper is organised as follows. In Section \ref{sec:method} we show the proposed methodology on how to train and generate Arabidopsis plants. In Section \ref{sec:exp} we report the results of our experiments. Finally, Section \ref{sec:conclusion} concludes the paper.


\begin{figure*}[t]
\includegraphics[width=0.9\textwidth]{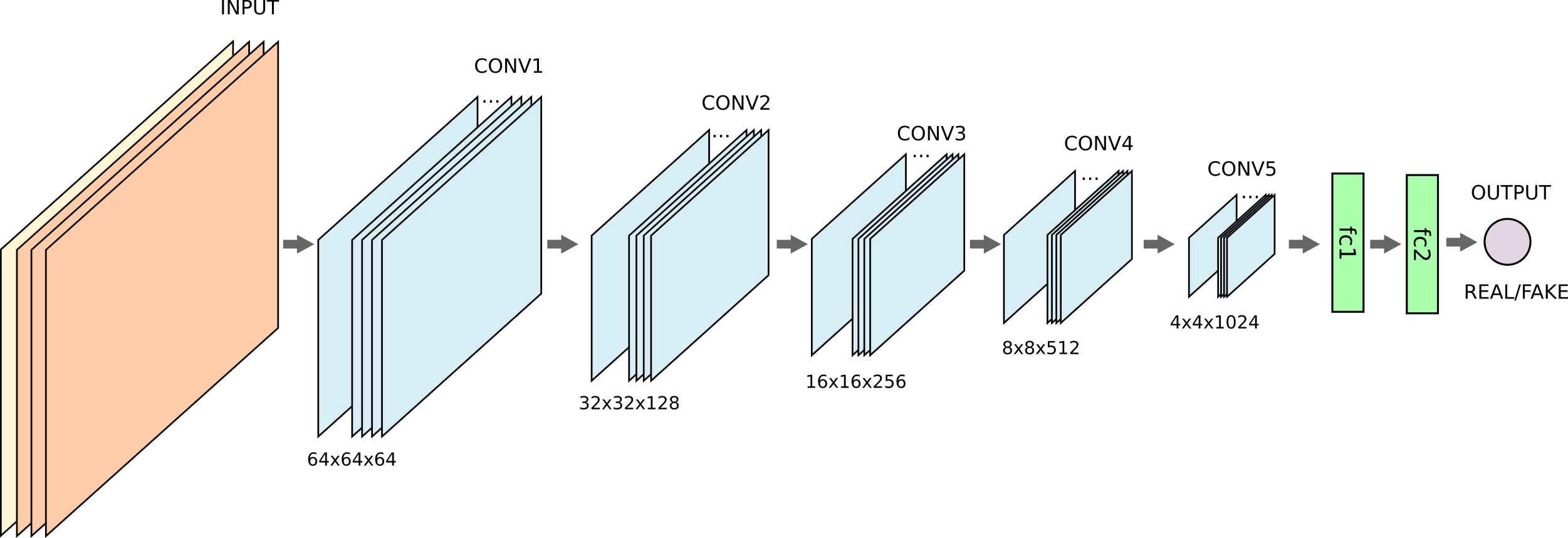}
\caption{\textbf{Discriminator.} This network takes as input an RGB image concatenated with the condition vector $y$ properly reshaped to be stacked as an additional channel. The rest of the network is a reversed version of $G$ (c.f. \figurename~\ref{fig:G}). The last node of the network is a binary classifier that discriminates between real and generated (fake) images.}
\label{fig:D}
\end{figure*}

\section{Methodology}
\label{sec:method}

To generate images of Arabidopsis plants, we followed the DCGAN architecture \cite{Radford2015}, but have added an extra (de)convolution layer to generate $128\times 128$ images. In Figures~\ref{fig:G} and~\ref{fig:D}, we show the generator and discriminator model respectively. Both networks share the same layer structure in reverse order. We provide further details in the next sections.

\subsection{Generative Adversarial Network}
\label{sec:gan}

A generative adversarial network has two models that train simultaneously: the generator $G$ and the discriminator $D$. The generator network takes as input a random vector $z \sim p_z(z)$ and learns the set of parameters $\theta_g$ to generate images $G(z;\theta_g)$ that follow the distribution of  real training images. At the same time, the discriminator $D$ learns a set of parameters $\theta_d$ to classify $x \sim p(x)$ as real images and $G(z;\theta_g)$ as synthetic (or fake) images. The training process maximises the probability of $D$ to assign the correct classes to $x$ and $G(z)$, whilst $G$ is trained to minimise $1-D(G(z;\theta_g))$. Using the cross-entropy as loss function, the objective $V(D,G)$ to be optimised is defined as follows:
\begin{equation}
    \begin{split}
    \min_G \max_D V(G,D) & = \mathbb{E}_{x\sim p_{data}(x)}\left[\log D(x)\right] \\ 
    &+ \mathbb{E}_{z\sim p_z(z)}\left[\log (1-D(G(z)))\right].
    \end{split}
    \label{eq:gan}
\end{equation}
The optimisation of \eqref{eq:gan} can be done via stochastic gradient descent, alternating the update of $\theta_d$ and $\theta_g$. 

In order to control the image to generate, we add $y \sim p_y(y)$ as input that embeds the condition \cite{Mirza2014}. Therefore, we have a set of real data (e.g., training set)
$\mathcal{D}_r = \left\lbrace (x_i,y_i) \right\rbrace^n_{i=1}$ for the discriminator and a set of sampled data $\mathcal{D}_s = \left\lbrace (z_i,y_i) \right\rbrace^n_{i=1}$ to train the generator. Hence, we update \eqref{eq:gan}, such that $D(x;\theta_d)$ becomes $D(x|y;\theta_d)$ and  $G(z;\theta_g)$ becomes $G(z|y;\theta_g)$.


The two networks, the generator and discriminator, could be networks of any architecture. In the next sections we provide details about $G$ and $D$. We used convolutional deep networks to generate images of Arabidopsis plants.

\subsection{The model \textit{G}}

Our model is inspired by \cite{Radford2015}, although we also added an additional (de)convolutional layer to obtain $128 \times 128$ images. The original model generates $64\times 64$ images which is not suitable for Arabidopsis plants synthesis, where young plants might be only a few pixels in size and mostly indistinguishable. 

The input layer takes a random variable $z \sim \mathbb{U}\left[-1,1\right]$ concatenated to a variable $y$ that sets the condition on the number of leaves. A typical approach for the condition is to use a one-hot encoding over the number of classes. We followed this approach, by considering the number of leaves as a \textit{category} on which a condition should be set, where $C$ denotes the number classes. 
Hence, a vector $y \in \left\lbrace 0,1 \right\rbrace^C$ will have all zeros, except for a `1' located at the position corresponding to a certain class of plants.  The condition $y$ within the training set $\mathcal{D}_r$ corresponds with the ground-truth leaf count, whereas the $y$ in $\mathcal{D}_s$ is randomly sampled, such that $y_t = 1$, where $t\sim \mathbb{U}[1,C]$ (namely, the `1' is located in a random location and the rest of the vector is filled with zeroes).

The so-formed input is then provided to two fully connected layers, denoted as \textit{fc1} and \textit{fc2}. The output of \textit{fc2} matches the size of the filters for the \textit{deconv1} layer, such that the output of the last fully connected layer can be easily reshaped. After 5 deconvolution layers, a $128\times 128\times 3$ output layer with \textit{tanh} activation function will present the generated plant image. We do not employ any upsampling, but we use $(2,2)$ stride instead, such that the network learns how to properly upscale between two consecutive deconvolutional layers. We adopted $5\times 5$ filter size on all the deconvolutional layers \cite{Radford2015}. Furthermore, the output of each layer is normalised \cite{Ioffe15} and passed through ReLU nonlinearity, before to be provided to the next layer. Similar setups also hold for the discriminator model. Although not graphically reported in \figurename~\ref{fig:G}, the condition $y$ is concatenated throughout all the steps of the network. In fact, each output of the fully connected layers has the vector $y$ added.\  The deconvolutional layers also have the (leaf count) conditions as additional feature maps, spatially replicating $y$ to properly match the layer size.  

\subsection{The model \textit{D}}

\figurename~\ref{fig:D} visualises the discriminator model. It can be seen as a inverted version of the generator, where the order of the layers is flipped and deconvolutional layers are replaced with convolutional ones. Also for this model, as discussed in Section~\ref{sec:gan}, the condition $y$ is embedded at all stages of the network. Here, the last layer of the network is a single node that outputs a binary value (fake vs. real images), activated with a \textit{sigmoid} function. Differently than the \textit{G}, the discriminator uses \textit{Leaky ReLU} \cite{Maas2013} as nonlinearity at each layer of the network \cite{Radford2015}, which has been shown to provide better convergence for classification. 

\section{Experimental results}
\label{sec:exp}
In this section we show the experimental results of training the ARIGAN. We implemented the network on Theano  \cite{Bastien2012} and the training was done on a  NVidia Tesla K8 GPU. Training takes $\sim10$ minutes per epoch on our setup.

\subsection{Dataset}

\begin{figure}[t]
\includegraphics[width=\linewidth]{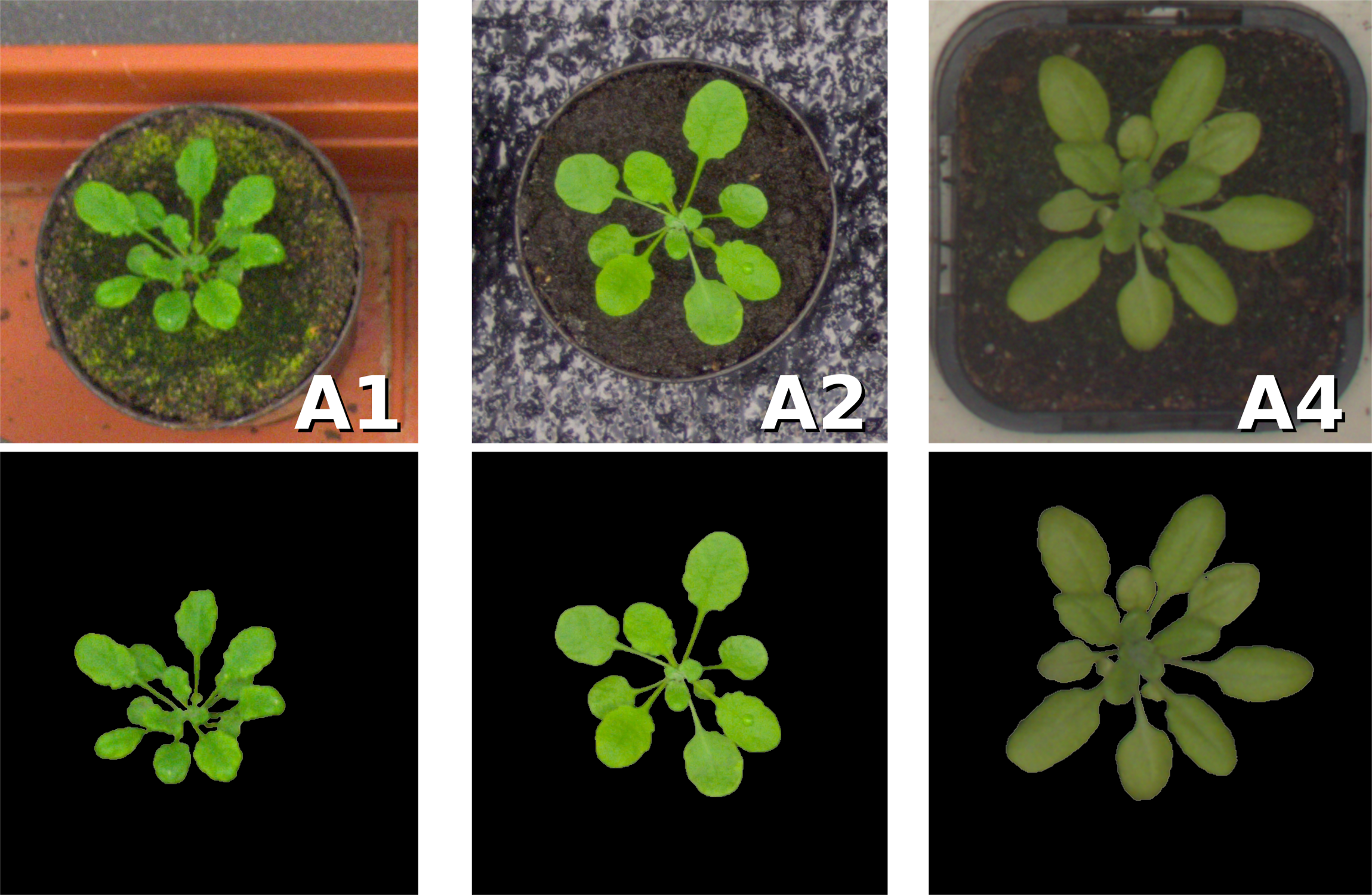}
\caption{\textbf{CVPPP Dataset.} Images used to train ARIGAN. We used the segmentation mask to relax the learning process, due to the significant variability of these setups.}
\label{fig:dataset}
\end{figure}

We used the CVPPP LCC 2017 plant dataset to train our model. These plant images are taken from different publicly available datasets \cite{Bell2016,Minervini2016}, containing Arabidopsis (A1, A2, and A4) and Tobacco (A3). Specifically, the training annotated datasets are:

\begin{enumerate}
    \item \textbf{A1:} 128 Arabidopsis Thaliana Col-0;
    \item \textbf{A2:} 31 Arabidopsis Thaliana of 5 differents cultivars;
    \item \textbf{A3:} 27 Tobacco plants;
    \item \textbf{A4:} 624 Arabidopsis Thaliana Col-0.
\end{enumerate}

For our purposes, we did not use A3 (Tobacco) dataset, due to the restricted number of data (27), compared to the Arabidopsis plants. Hence, we trained our network with the A1+A2+A4 dataset (c.f. \figurename~\ref{fig:dataset}), containing a total of 783 images. Even though this number of training images is much bigger than in the previous CVPPP 2015 challenge\footnote{The A4 dataset was not included.}, it is still low for training a deep neural network with many parameters.\footnote{Typically GANs are trained with data in order of magnitude of thousands or millions images \cite{Radford2015}.} To overcome this issue, we performed dataset augmentation, by rotating the images by ten equidistant angles of the range $[0,2\pi)$, we also applied horizontal and vertical flipping to further increase variability, obtaining an overall 30-fold increase in the number of training images.

Input images were pre-processed to be all in the same size of $128\times 128$ by cropping  (to be made square) and rescaling. In order to match the output values of the \textit{tanh}, we mapped the range of values from $[0,255]$ to $[-1,1]$.

\begin{figure*}
\includegraphics[width=\textwidth]{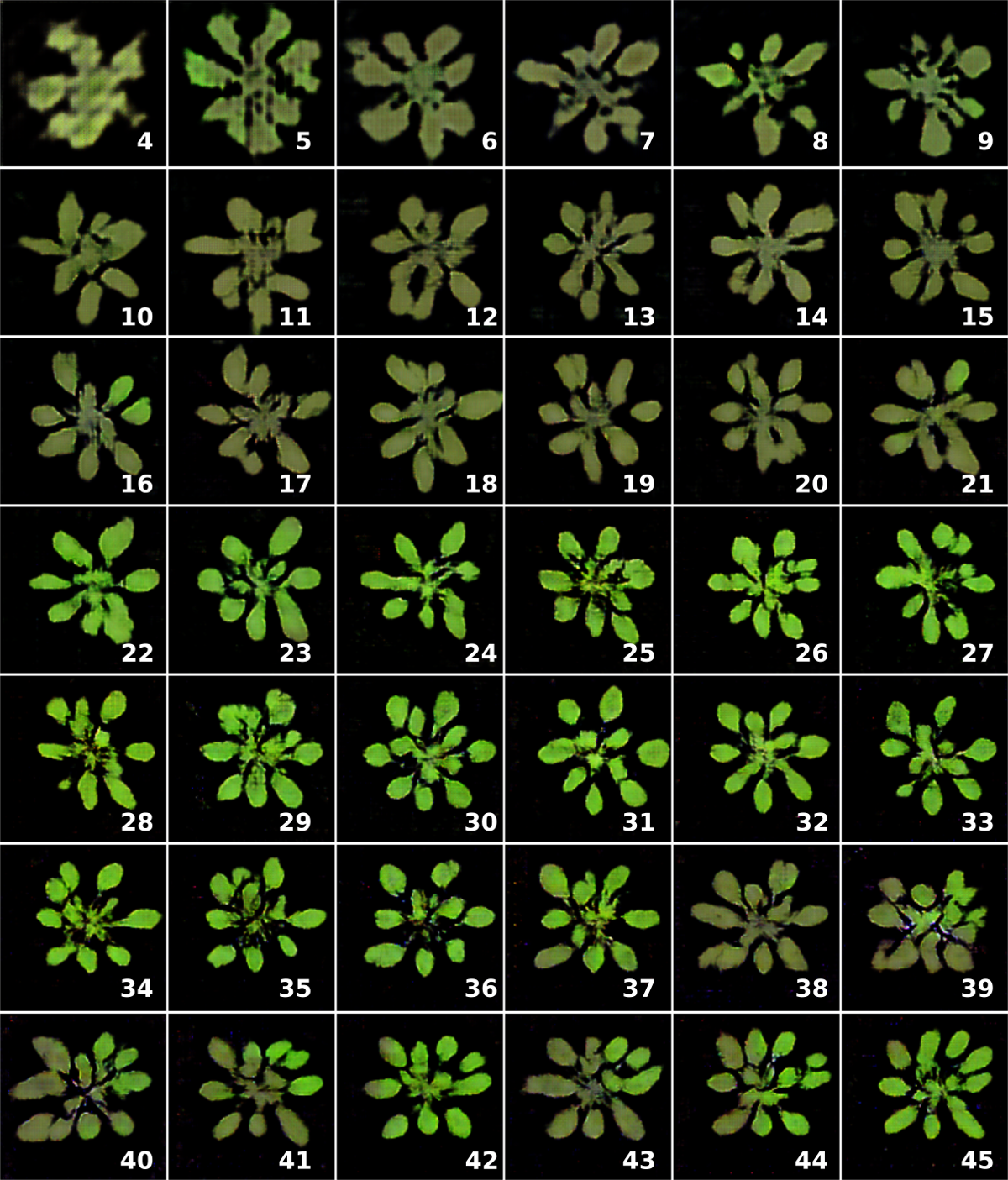}
\caption{\textbf{Generated plants.} Fixed sample noise $z$ is provided to the generator during training of ARIGAN. The number reported in the bottom right corner of each image refers to the epoch number. (\textit{Best viewed in colour}.)}
\label{fig:panel}
\end{figure*}


\subsection{Qualitative results}

In \figurename~\ref{fig:panel}, we show generated plants at different training epochs. Specifically, we sampled a single random input along with a random condition and we gave it to the generator network (c.f. \figurename~\ref{fig:G}). It can be seen that a clear Arabidopsis plant is obtained in about 30 epochs. In can be observed that some images have a light green appearance (typical of A1 and A2 dataset), others have dark green (A4), and some others have a mixture of those. Lastly, we see that at some epochs, the synthesised image contains a mixture of light and dark green texture, blended together seamlessly. However, generated images lack high frequencies, causing absence of some details, e.g., leaf veins or petiole.

Encouraged by these results, we extracted a subset of images to create a new dataset. We do this to quantitatively evaluate this artificial data using a state-of-the-art algorithm for leaf counting \cite{Giuffrida2015}.

\subsection{The \textit{Ax} Dataset}

\begin{figure}[t]
\includegraphics[width=\linewidth]{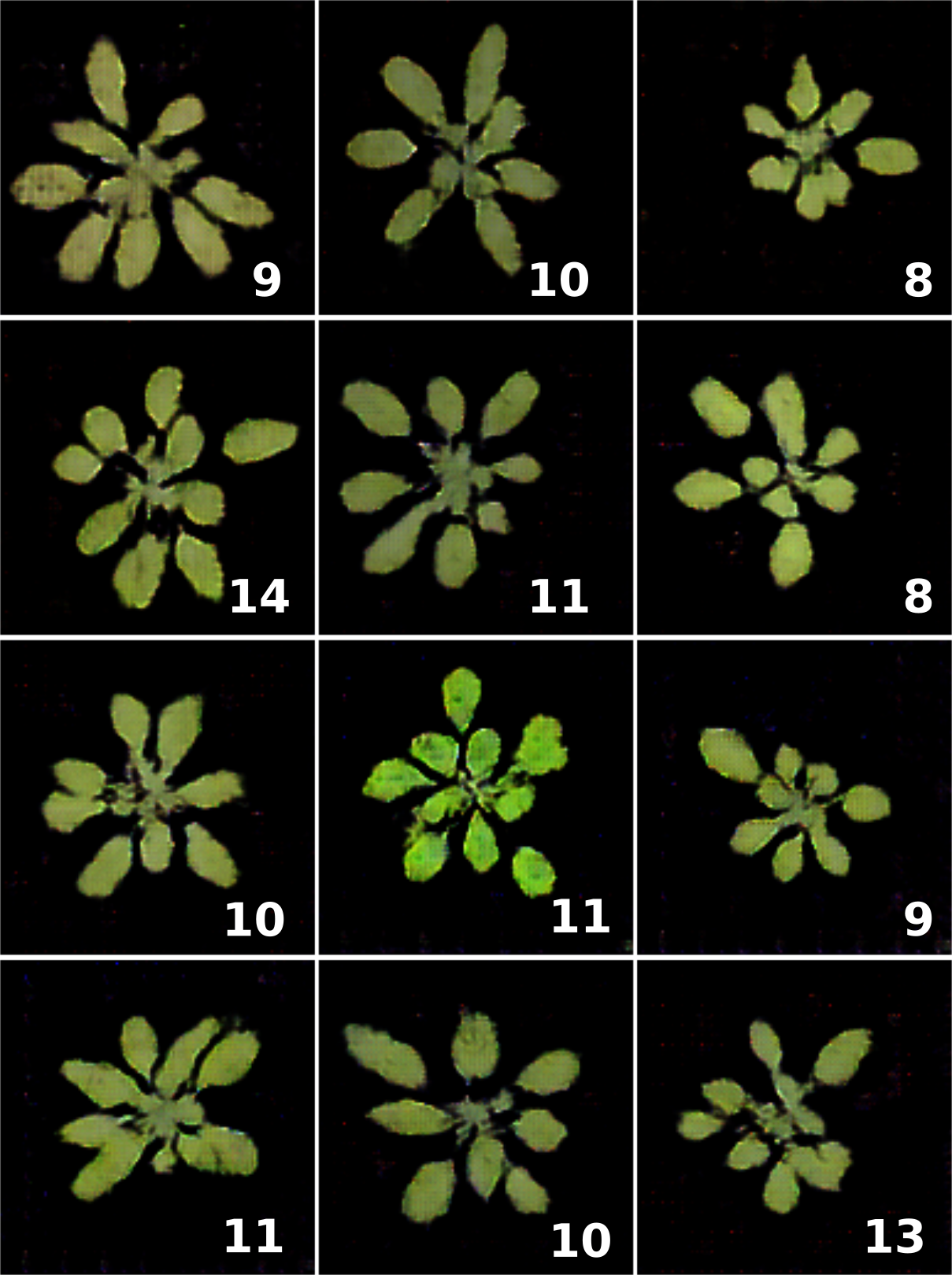}
\caption{\textbf{Ax Dataset.} Samples from the Ax dataset generated with ARIGAN. Bottom-right numbers refer to the leaf count. (\textit{Best viewed in colour}).}
\label{fig:ax}
\end{figure}

Using our model, we artificially generated a new dataset of images. Following the nomenclature policy used for the CVPPP workshop, we decided to call it \textit{Ax}. Specifically, we collected 57 images from our model some samples of which are shown in \figurename~\ref{fig:ax}. We collected the images at
different stages of learning (we trained at least 20 epochs), providing random noise and conditions. Then we named the files using the \textit{plantXXX\_rgb.png} format, listing them in a CSV file with the count of leaves. We did not provide plant segmentation masks, as our algorithm does not generate background.

\subsection{Quantitative results}

We evaluated the Ax dataset to train the leaf counting algorithm in \cite{Giuffrida2015}. For our experiments, we used the A4 dataset, as it contains the most number of images (624). The evaluation was done using 4-fold cross validation, where 468 images are randomly selected for training, and the remaining 156 for testing. We kept SVR parameters at their standard values, except $C=3$. (Please refer to \cite{Drucker1997} for further details about the parameter $C$).


In Table~\ref{tab:qres} we show the results of these experiments, using the evaluation metrics employed in \cite{Minervini2017}. Specifically, the table reports the results using A4 dataset only (top set of lines), paired with the results using Ax from training as well.\footnote{Ax images were added as part of the training set. Specifically, for each split of the cross validation, 57 images were added to make a training set of 525 plants images.} Overall, considering that the leaf counting algorithm in \cite{Giuffrida2015} has not been demonstrated to work when different training sets are provided, we found that the additional dataset Ax improves the testing errors and reduces overfitting.


\begin{table}[t]
\begin{center}
\begin{tabular}{@{}rll@{}}
\toprule
 & ~\textbf{Training Error}~ & ~\textbf{Testing Error}~ \\ \midrule
 \multicolumn{3}{c}{\textit{Trained on A4 only}} \\
\multicolumn{1}{r}{DiC}   & 0.013 (0.185)             & 0.147 (1.362)             \\
\multicolumn{1}{r}{$|$DiC$|$}   & 0.026 (0.183)       & 0.942 (0.992)             \\
\multicolumn{1}{r}{MSE}     & 0.031              & 1.865             \\
\multicolumn{1}{r}{R$^2$} & 0.999              & 0.947
\\ \midrule \midrule
 \multicolumn{3}{c}{\textit{Trained on A4 and Ax}} \\
\multicolumn{1}{r}{DiC}        & 0.229 (0.370)  & 0.186 (1.253)             \\
\multicolumn{1}{r}{$|$DiC$|$}  & 0.042 (0.368)  & 0.891 (0.899)             \\
\multicolumn{1}{r}{MSE}        & 0.137              & 1.596             \\
\multicolumn{1}{r}{R$^2$}      & 0.996             & 0.955             \\ \bottomrule
\end{tabular}
\end{center}
\caption{\textbf{Quantitative results.} We trained the leaf counting algorithm in \cite{Giuffrida2015} using A4 dataset only (top set of lines) and A4+Ax (bottom set of lines). Results obtained with 4-fold cross validation. Results for DiC and $|$DiC$|$ are reported as \textit{mean} (\textit{std}).}
\label{tab:qres}
\end{table}

\section{Conclusions}
\label{sec:conclusion}

Image-based plant phenotyping has received a great interest from the machine learning community in the last years. In fact, different methodologies have been proposed for leaf segmentation \cite{Ren2016,RomeraParedes2016,Scharr2016} and leaf counting \cite{Giuffrida2015,Pape2015}. Despite the recent release of several plant phenotyping datasets \cite{Bell2016,Cruz2016,Minervini2016}, the quantity of data is still the main issue for machine learning algorithm \cite{Tsaftaris2016}, especially for those using a vast number of parameters (e.g., deep network approaches).

In this paper, in other to alleviate the lack of training data in plant phenotyping, we use a generative model to create  synthetic Arabidopsis plants. Recently, Generative Adversarial Networks \cite{Goodfellow2014} were proposed, which have been proved to create realistic natural images. Encouraged from their results, we wanted to train a GAN in order to generate  plants. Using the CVPPP dataset (only A1, A2, and A4), we trained an adversarial network inspired by DCGAN \cite{Radford2015} to generate synthetic Arabidopsis images. Our \textit{Arabidopsis Rosette Image Generator (through) Adversarial Network} (ARIGAN) is able to produce realistic $128\times 128$ colour images of plant. Specifically, our network falls into the category of \textit{Conditional GAN}, where an additional input of the network allows to set a condition over the number of leaves of a plant.

From our experiments, we found that ARIGAN learns how to generate realistic images of plant after a few of iterations (c.f. \figurename~\ref{fig:panel}) . This qualitative results led us to create a dataset of artificial Arabidopsis plants images. Therefore, we gathered 57 images that our network generated to make the \textit{Ax} dataset, as displayed in \figurename~\ref{fig:ax}. We evaluated our synthetic dataset using to train a state-of-the-art leaf counting algorithm \cite{Giuffrida2015}. Our quantitative experiments show that the extension of the training dataset with the images in Ax improved the testing error and reduced overfitting. We run a 4-fold cross validation experiment on A4 dataset. Evaluation metrics of our experiments are reported in Table \ref{tab:qres}. Our synthetic dataset \textit{Ax} is available to download at \url{http://www.valeriogiuffrida.academy/ax}.

\section*{Acknowledgements}
\noindent
This work was supported by The Alan Turing Institute under the EPSRC grant EP/N510129/1, and also by the BBSRC grant BB/P023487/1.

{\small
\bibliographystyle{ieee}
\bibliography{eg_paper_for_arxiv}

\begin{thebibliography}{10}\itemsep=-1pt

\bibitem{Bastien2012}
F.~Bastien, P.~Lamblin, R.~Pascanu, J.~Bergstra, I.~J. Goodfellow, A.~Bergeron,
  N.~Bouchard, and Y.~Bengio.
\newblock Theano: new features and speed improvements.
\newblock Deep Learning and Unsupervised Feature Learning Workshop (NIPS),
  2012.

\bibitem{Bell2016}
J.~Bell and H.~Dee.
\newblock {Aberystwyth Leaf Evaluation Dataset}, 2016.

\bibitem{Cappelli2000}
R.~Cappelli, A.~Erol, D.~Maio, and D.~Maltoni.
\newblock Synthetic fingerprint-image generation.
\newblock In {\em International Conference on Pattern Recognition.}, volume~3,
  pages 471--474 vol.3, 2000.

\bibitem{Cruz2016}
J.~A. Cruz, X.~Yin, X.~Liu, S.~M. Imran, D.~D. Morris, D.~M. Kramer, and
  J.~Chen.
\newblock {Multi-modality imagery database for plant phenotyping}.
\newblock {\em Machine Vision and Applications}, 27(5):735--749, 2016.

\bibitem{Denton2015}
E.~Denton, S.~Chintala, A.~Szlam, and R.~Fergus.
\newblock {Deep Generative Image Models using a Laplacian Pyramid of
  Adversarial Networks}.
\newblock In C.~Cortes, N.~D. Lawrence, D.~D. Lee, M.~Sugiyama, and R.~Garnett,
  editors, {\em Advances in Neural Information Processing Systems 28}, pages
  1486--1494. Curran Associates, Inc., 2015.

\bibitem{DiPaola2009}
S.~DiPaola and L.~Gabora.
\newblock {Incorporating characteristics of human creativity into an
  evolutionary art algorithm}.
\newblock {\em Genetic Programming and Evolvable Machines}, 10(2):97--110,
  2009.

\bibitem{Dosovitskiy2015}
A.~Dosovitskiy, J.~T. Springenberg, and T.~Brox.
\newblock Learning to generate chairs with convolutional neural networks.
\newblock In {\em 2015 IEEE Conference on Computer Vision and Pattern
  Recognition (CVPR)}, pages 1538--1546, 2015.

\bibitem{Drucker1997}
H.~Drucker, C.~J.~C. Burges, L.~Kaufman, A.~J. Smola, and V.~Vapnik.
\newblock {Support Vector Regression Machines}.
\newblock In M.~C. Mozer, M.~I. Jordan, and T.~Petsche, editors, {\em Advances
  in Neural Information Processing Systems 9}, pages 155--161. MIT Press, 1997.

\bibitem{Giuffrida2015}
M.~V. Giuffrida, M.~Minervini, and S.~Tsaftaris.
\newblock {Learning to Count Leaves in Rosette Plants}.
\newblock In {\em CVPPP workshop - BMVC}, page~13. British Machine Vision
  Association, 2015.

\bibitem{Goodfellow2014}
I.~J. Goodfellow, J.~Pouget-Abadie, M.~Mirza, B.~Xu, D.~Warde-Farley, S.~Ozair,
  A.~Courville, and Y.~Bengio.
\newblock {Generative Adversarial Networks}.
\newblock pages 1--9, 2014.

\bibitem{Gregor2015}
K.~Gregor, I.~Danihelka, A.~Graves, D.~J. Rezende, and D.~Wierstra.
\newblock {DRAW: A Recurrent Neural Network For Image Generation}.
\newblock 2015.

\bibitem{Hochreiter1997}
S.~Hochreiter and J.~Schmidhuber.
\newblock {Long Short-Term Memory}.
\newblock {\em Neural Computation}, 9(8):1735--1780, 1997.

\bibitem{Ioffe15}
S.~Ioffe and C.~Szegedy.
\newblock Batch normalization: Accelerating deep network training by reducing
  internal covariate shift.
\newblock In F.~Bach and D.~Blei, editors, {\em Proceedings of the 32nd
  International Conference on Machine Learning}, volume~37 of {\em Proceedings
  of Machine Learning Research}, pages 448--456, Lille, France, 07--09 Jul
  2015. PMLR.

\bibitem{mnist}
Y.~LeCunn.
\newblock {The MNIST database of handwritten digits,
  \url{http://yann.lecun.com/exdb/mnist/}}.

\bibitem{Maas2013}
A.~L. Maas, A.~Y. Hannun, and A.~Y. Ng.
\newblock Rectifier nonlinearities improve neural network acoustic models.
\newblock In {\em in ICML Workshop on Deep Learning for Audio, Speech and
  Language Processing}, 2013.

\bibitem{Minervini2016}
M.~Minervini, A.~Fischbach, H.~Scharr, and S.~A. Tsaftaris.
\newblock {Finely-grained annotated datasets for image-based plant
  phenotyping}.
\newblock {\em Pattern Recognition Letters}, 81:80--89, 2016.

\bibitem{Minervini2017}
M.~Minervini, M.~V. Giuffrida, P.~Perata, and S.~A. Tsaftaris.
\newblock {Phenotiki: An open software and hardware platform for affordable and
  easy image-based phenotyping of rosette-shaped plants.}
\newblock {\em The Plant journal: for cell and molecular biology}, 2017.

\bibitem{Mirza2014}
M.~Mirza and S.~Osindero.
\newblock {Conditional Generative Adversarial Nets}.
\newblock In {\em arXiv}, 2014.

\bibitem{Muendermann2005}
L.~M{\"u}ndermann, Y.~Erasmus, B.~Lane, E.~Coen, and P.~Prusinkiewicz.
\newblock Quantitative modeling of arabidopsis development.
\newblock {\em Plant Physiology}, 139(2):960--968, 2005.

\bibitem{Pape2015}
J.-M. Pape and C.~Klukas.
\newblock {Utilizing machine learning approaches to improve the prediction of
  leaf counts and individual leaf segmentation of rosette plant images}.
\newblock {\em Proceedings of the Computer Vision Problems in Plant Phenotyping
  (CVPPP)}, pages 1--12, 2015.

\bibitem{Radford2015}
A.~Radford, L.~Metz, and S.~Chintala.
\newblock {Unsupervised Representation Learning with Deep Convolutional
  Generative Adversarial Networks}.
\newblock 2015.

\bibitem{Ren2016}
M.~Ren and R.~S. Zemel.
\newblock {End-to-End Instance Segmentation and Counting with Recurrent
  Attention}.
\newblock 2016.

\bibitem{RomeraParedes2016}
B.~Romera-Paredes and P.~H.~S. Torr.
\newblock {Recurrent Instance Segmentation}.
\newblock In B.~Leibe, J.~Matas, N.~Sebe, and M.~Welling, editors, {\em
  Computer Vision -- ECCV 2016: 14th European Conference, Amsterdam, The
  Netherlands, October 11-14, 2016, Proceedings, Part VI}, pages 312--329.
  Springer International Publishing, Cham, 2016.

\bibitem{Scharr2016}
H.~Scharr, M.~Minervini, A.~P. French, C.~Klukas, D.~M. Kramer, X.~Liu,
  I.~Luengo, J.-M. Pape, G.~Polder, D.~Vukadinovic, X.~Yin, and S.~A.
  Tsaftaris.
\newblock {Leaf segmentation in plant phenotyping: a collation study}.
\newblock {\em Machine Vision and Applications}, 27(4):585--606, 2016.

\bibitem{Tsaftaris2016}
S.~A. Tsaftaris, M.~Minervini, and H.~Scharr.
\newblock {Machine Learning for Plant Phenotyping Needs Image Processing}.
\newblock {\em Trends in Plant Science}, 21(12):989--991, dec 2016.

\bibitem{Han2016}
H.~Zhang, T.~Xu, H.~Li, S.~Zhang, X.~Huang, X.~Wang, and D.~Metaxas.
\newblock Stack{GAN}: Text to photo-realistic image synthesis with stacked
  generative adversarial networks.
\newblock {\em arXiv}, 2016.

\end{thebibliography}
}

\end{document}